\documentclass[letterpaper,10pt,conference]{ieeeconf/ieeeconf}

\IEEEoverridecommandlockouts 

\usepackage{amsmath}
\usepackage{amssymb}
\usepackage{booktabs}
\usepackage{balance}
\usepackage{graphicx}
\usepackage{xcolor}
\usepackage{url}
\usepackage{tikz}
\usetikzlibrary{arrows.meta}
\graphicspath{{ieeeconf/}}

\title{\LARGE \bf
Stable and Efficient Real-World Online VLA Post-Training via\\
Asynchronous Replay-Anchored Policy Improvement
}

\author{Jiarui Yang$^{1,2,*}$, Jiajin Zhang$^{1,2,*}$, Bin Zhu$^{3}$,
Jingjing Chen$^{1,2,\dagger}$, and Yu-Gang Jiang$^{1,2}$%
\thanks{$^{*}$Equal contribution. $^{\dagger}$Corresponding author.
$^{1}$Institute of Trustworthy Embodied AI, Fudan University;
$^{2}$Shanghai Key Laboratory of Multimodal Embodied AI;
$^{3}$Singapore Management University. Email: \texttt{jryang24@m.fudan.edu.cn}.}%
}

\makeatletter
\IEEEaftertitletext{%
  \begin{minipage}{\textwidth}
    \centering
    \includegraphics[width=\textwidth]{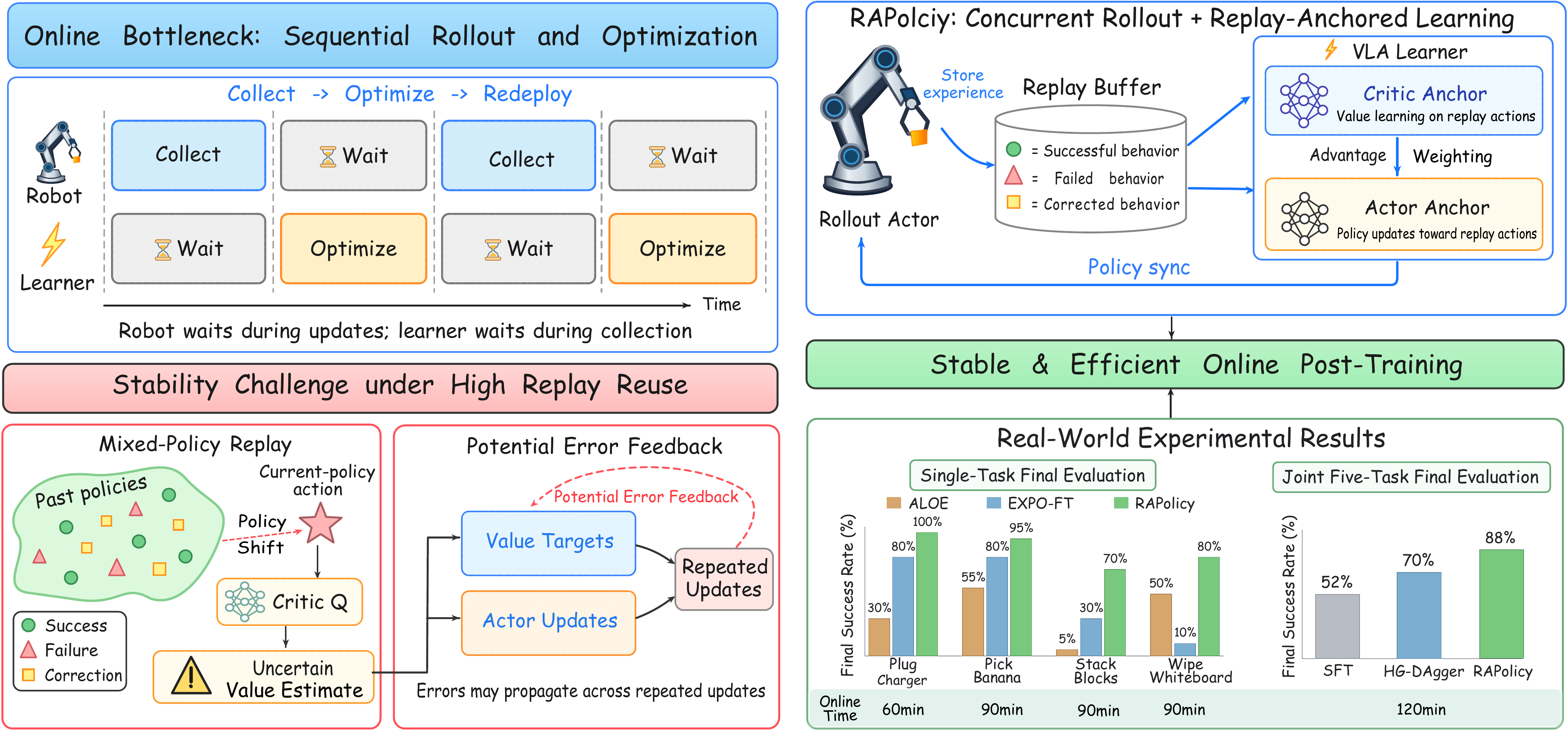}
    \def\@captype{figure}%
    \caption{Motivation and overview of RAPolicy. Left: sequential rollout and
    optimization introduce waiting, while evaluating actions outside replay
    coverage can propagate value errors. Right: RAPolicy combines asynchronous
    execution with replay-anchored critic and actor learning. Under the same
    online training budgets, RAPolicy achieves substantially higher overall task
    success rates than the evaluated baselines.}
    \label{fig:intro_motivation}
  \end{minipage}
}
\makeatother

\begin{document}

\maketitle
\thispagestyle{empty}
\pagestyle{empty}

\begin{abstract}
Online post-training of vision-language-action (VLA) models requires efficient
use of robot interaction and reliable policy improvement from continually
collected experience. We propose asynchronous Replay-Anchored Policy
improvement (RAPolicy), a framework that performs rollout and learning concurrently
while grounding both critic and actor updates in replayed behavior. The critic
learns chunk-level values from recorded actions and constructs Bellman targets
without predicting next actions, reducing computation and dependence on
action-value estimates outside replay coverage. The one-step flow actor reuses
the initial noise stored during rollout and learns through advantage-weighted
conditional likelihood, directly supervising the action mapping used for
execution. We evaluate RAPolicy across four single-task settings and one joint
five-task setting in the real world, with online training budgets of only 1--2 hours.
Starting from policies fine-tuned on just 10 demonstrations per task, RAPolicy
rapidly adapts to new single tasks and achieves an average 86.3\% success
rate. In the joint multi-task setting, RAPolicy improves overall success rate
from 52\% to 88\% while preserving performance on already reliable tasks and
improving weaker capabilities. Overall, RAPolicy substantially outperforms the baselines in aggregate task success while requiring fewer human
interventions, demonstrating stable policy improvement and high online
training efficiency. Project page: \url{https://flyfaerss.github.io/RAPolicy/}.
\end{abstract}

\section{INTRODUCTION}
\label{sec:introduction}

Vision-language-action (VLA) models acquire broad manipulation capabilities
from large and heterogeneous pretraining corpora~\cite{brohan2023rt2,black2024pi0,black2025pi05}. Yet a
pretrained checkpoint can remain unreliable on precision-critical deployment
tasks, where small errors in an action chunk may cause contact failure or
require human intervention. Physical deployment also produces exactly the
task-specific evidence needed for improvement: successful executions,
unsuccessful attempts, safety stops, and corrective human actions. Online
post-training offers a way to convert this experience into improved
behavior~\cite{guo2025irevla,amin2025recap,xu2026rltoken,yang2026structrl}.
Our goal is continuous deployment-time adaptation: as the robot continues to
gather experience, its VLA should rapidly learn from the collected data,
promptly incorporating lessons from failures and human-intervention
trajectories to complete tasks more reliably with less human assistance.

Achieving this goal efficiently calls for reducing waiting between interaction
and optimization while learning effectively from deployment experience. Physical
interaction consumes robot time, while updating the VLA action policy is
computationally intensive. A stage-wise online loop collects a batch of
experience, optimizes the policy between collection rounds, and then deploys
the updated policy~\cite{dong2026batchonline,yang2026aloe}. This
serializes two costly activities and can leave the robot idle during training
and the learner idle during collection. Overlapping rollout and
learning offers a way to reduce this waiting~\cite{nair2015gorila,espeholt2018impala,horgan2018apex}, but more concurrent computation
does not by itself imply faster policy improvement.

The replay buffer contains experience generated by different policy versions,
including exploratory behavior, successes, failures, and human corrections. In
an asynchronous system, the robot continues collecting data with the most
recently synchronized policy while the learner updates the current policy.
The deployed policy can therefore differ from the learner's policy between
synchronizations. Continuous online adaptation requires a learning method that
can stably reuse this heterogeneous experience from earlier policies while
avoiding repeated policy updates toward actions whose values are poorly
supported by the collected data.

Existing learning strategies expose this tension. Behavior cloning learns
recorded actions but does not by itself perform return-based credit
assignment~\cite{zhao2023act,chi2023diffusionpolicy}.
Success-only fine-tuning uses outcomes to select trajectories, but discards
failures rather than using them for value learning. Actor--critic learning can
propagate outcomes through replay, yet current-policy action bootstrapping or
direct critic action gradients can query newly generated actions with weak
data coverage~\cite{kumar2020cql,kostrikov2022iql}. Errors at those actions can affect subsequent value targets or
actor updates. This is a potential feedback mechanism, not an inevitable
failure of off-policy learning; it motivates grounding both sides of policy
improvement in recorded experience.

Building on these considerations, we propose asynchronous Replay-Anchored Policy improvement (RAPolicy), which combines concurrent execution and
behavior-grounded learning for continuous online VLA adaptation. RAPolicy
overlaps rollout and optimization while grounding critic and actor updates
in recorded behavior---a design we call \emph{replay anchoring}. The critic
constructs chunk-level Bellman targets without sampling next actions from
the evolving policy, while the actor learns replayed action chunks through
advantage-weighted conditional likelihood without critic action gradients.
This reduces reliance on poorly supported actions while retaining failures
for value learning. To support efficient rollout and optimization, RAPolicy uses
the VLA's one-step flow actor, with stored rollout latents conditioning replay
updates and no multi-step generation chain to unfold for this objective.

We evaluate RAPolicy in the real world across four single-task
and one joint five-task setting. The single-task experiments cover precise
manipulation in Plug Charger and Stack Blocks, contact-rich manipulation
and visual understanding in Wipe Whiteboard, and pick-and-place manipulation
in Pick Banana. Under the same online-time budget for each task, including
both rollout and training, RAPolicy achieves a 72.5\% relative improvement in
average policy success rate over EXPO-FT, demonstrating higher online
adaptation efficiency. In the joint five-task setting, RAPolicy improves
post-adaptation performance over HG-DAgger, maintaining strong success rates
on easier tasks while improving success on harder tasks. These results
demonstrate the effectiveness of RAPolicy and its potential for multi-task
online adaptation.

Our main contributions are:
\begin{itemize}
    \item We present RAPolicy, an online VLA post-training framework that
    combines asynchronous rollout and optimization with replay-anchored
    learning to support stable and efficient deployment-time adaptation.

    \item We develop a replay-anchored update scheme that couples
    chunk-level value learning with advantage-weighted conditional
    likelihood optimization of a one-step VLA actor. Recorded actions
    and rollout latents ground policy improvement in replay, without
    current-policy next-action sampling or critic action gradients.

    \item We demonstrate the effectiveness of RAPolicy through real-world
    experiments across four single-task and joint five-task settings, showing higher policy success rates under matched per-task
    online-time budgets and gains on challenging tasks during
    multi-task adaptation.
\end{itemize}

\section{RELATED WORK}
\label{sec:related_work}

\subsection{Reinforcement Learning for VLA Post-Training}
Robot reinforcement learning improves policies using autonomous experience and human
corrections~\cite{kalashnikov2018qtopt,haarnoja2018sac,luo2024serl,luo2024hilserl}, and VLA post-training extends
this approach to pretrained generalist policies~\cite{amin2025recap,xu2026rltoken,mark2024parl,dong2026expoft}.
For continued deployment, a key distinction is whether collection and
optimization alternate or overlap. Batch-online RL and ALOE adopt
collect--then-update loops~\cite{dong2026batchonline,yang2026aloe}, whereas
asynchronous RL decouples acting and learning~\cite{nair2015gorila,espeholt2018impala,horgan2018apex}.
RAPolicy combines the latter execution principle with replay-anchored VLA
updates, addressing both waiting time and learning from accumulated experience.

\subsection{Replay-Based Policy Improvement for Generative Policies}
Replay-based RL limits exploitation of value errors through conservative
critics~\cite{kumar2020cql} or advantage-weighted behavior learning~\cite{kostrikov2022iql,haarnoja2018sac,nair2020awac,ball2023rlpd};
IQL additionally uses in-sample backups. For generative policies,
value information can guide sampling or flow learning~\cite{yang2026aloe},
while chunk-level critics account for temporally extended actions~\cite{li2025qchunking,seo2025cqnas}.
FQL combines Q maximization with distillation from a multi-step flow policy
into a separate one-step actor~\cite{park2025fql}. RAPolicy instead couples
IQL-style chunk-value backups with advantage-weighted conditional likelihood
updates of one-step actor, using stored rollout latents to condition
actor learning. This combination reduces reliance on value estimates for
actions outside replay coverage while keeping policy updates computationally
lightweight.

\section{METHOD}
\label{sec:method}

We consider online adaptation of a pretrained, language-conditioned VLA on
one physical robot. At decision $t$, the input $s_t=(o_t,x_t,\ell)$ comprises
images, robot state, and a task instruction. The deployed VLA proposes an
action chunk of horizon $H$, denoted $\mathbf a_t$;
a human operator may correct its execution. The resulting behavior, rewards,
and next observation provide online learning experience.

\subsection{Overview and Asynchronous Execution}
\begin{figure}[t]
    \centering
    \includegraphics[width=\columnwidth]{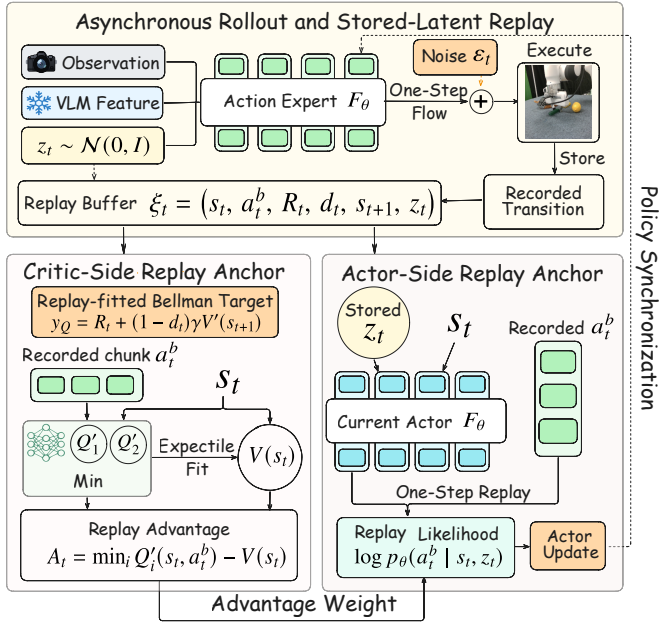}
    \caption{RAPolicy framework. Rollout and learning proceed concurrently.
    Replay-anchored value learning supplies Bellman targets and advantages,
    while the one-step actor reuses stored rollout latents for
    advantage-weighted conditional-likelihood updates. Updated actor parameters
    are synchronized with rollout after each update group.}
    \label{fig:method_update}
\end{figure}

RAPolicy combines asynchronous execution with replay-anchored learning to support
continuous VLA adaptation. The robot collects experience while the learner
updates the policy, reducing waiting between interaction and optimization.
Chunk-level value learning and advantage-weighted actor updates reuse recorded behavior without using Q-value estimates for actions generated by the current policy. The one-step flow actor reuses stored rollout latents, keeping action generation and replay updates computationally lightweight. An overview of RAPolicy is shown in Fig.~\ref{fig:method_update}.

We denote the learner policy parameters by $\theta$ and the rollout policy
parameters by $\bar\theta$, with actions sampled from
$\pi_{\bar\theta}(\mathbf a_t\mid s_t)$.
Rollout and optimization proceed concurrently, allowing the learner to update
from recent rollout data without pausing data collection. Each update group
uses a fixed replay sampling set;
rollout episodes completed during the group are added before the next one.
After each group, the rollout policy is synchronized with the learner and
remains fixed until the next synchronization.

In practice, each update group consists of 20 critic and 5 actor updates,
enabling frequent updates to the rollout policy. We also warm up the critic for
2,560 updates so that it can provide a reliable optimization signal for the
actor.

\subsection{Replay Construction}
We adopt the pretrained $\pi_{0.5}$~\cite{black2025pi05} as our actor.
Given state $s_t$, it predicts a normalized delta-pose action chunk
$\mathbf a_t=(a_{t,0},\ldots,a_{t,H-1})$ of horizon $H$. Human corrections are
mapped into the same normalized action space and incorporated into the recorded
behavior chunk $\mathbf a_t^b$, which is used by both the critic and actor.

Each transition is represented as $(s_t,\mathbf a_t^b,R_t,d_t,s_{t+1},z_t)$, where $R_t$ is
the cumulative discounted reward associated with executing the recorded chunk,
$d_t$ indicates termination, and $s_{t+1}$ is the state after its execution.
Here, $z_t$ is the initial Gaussian latent sampled by the flow actor when predicting the
rollout action chunk. It is stored with
the transition and reused to condition
subsequent actor updates.

We organize these transitions into an online buffer and a demo
buffer. The online buffer stores all transitions from completed rollout
episodes, whereas the demo buffer stores intervention transitions
and all transitions from the 20 fastest successful trajectories discovered
during rollout. As rollout discovers faster successful trajectories, they
replace slower ones in the demo buffer, providing continuously improved
demonstrations for policy optimization. Training batches are sampled from both
buffers at a 50:50 ratio, and the resulting sampling distribution is denoted
by $\mathcal D$.

\subsection{Replay-Anchored Chunk Value Learning}
We train the critic with in-sample value learning~\cite{kostrikov2022iql} over
the replay distribution $\mathcal D$, using only recorded behavior chunks to
anchor value learning to the replay data. A frozen VLM and a shared state
backbone encode each state, while a GRU encodes each action chunk. The value
head takes the state representation as input, whereas the twin Q heads take
both the state and action-chunk representations.

We denote the value and Q functions by $V$ and $Q_i$, and their EMA target
counterparts by $V'$ and $Q_i'$, respectively. For the recorded behavior chunk
$\mathbf a_t^b$, the Bellman target is
\begin{equation}
y_Q=R_t+(1-d_t)\gamma V'(s_{t+1}).
\label{eq:q_target}
\end{equation}
Here, $\gamma$ is the discount factor between consecutive action chunks, and
$(1-d_t)$ removes bootstrapping for terminal transitions.
Then, we optimize the twin Q-functions with
\begin{equation}
\mathcal L_Q=\frac{1}{2}\sum_{i=1}^2
\mathbb E_{\mathcal D}
\left[(Q_i(s_t,\mathbf a_t^b)-y_Q)^2\right].
\label{eq:q_loss}
\end{equation}
To fit $V(s_t)$ to the $\tau$-expectile of the Q-values of replayed behavior
chunks, we define
\[
\delta_t=\min_{i\in\{1,2\}}Q_i'(s_t,\mathbf a_t^b)-V(s_t)
\]
and minimize
\begin{equation}
\mathcal L_V=\mathbb E_{\mathcal D}
\left[w_\tau(\delta_t)\delta_t^2\right],
\label{eq:v_loss}
\end{equation}
where $w_\tau(\delta)=\tau$ if $\delta\geq0$ and $1-\tau$ otherwise.
In practice, we set $\tau=0.7$ to bias $V(s_t)$ toward the higher Q-values of
replayed behavior chunks.
This upper-expectile baseline allows the actor update to emphasize replayed
chunks whose Q-values exceed $V(s_t)$.
The target value and Q heads are updated by EMA after each critic step.

The value function is fitted solely to Q-values of replayed behavior chunks.
Therefore, the Bellman target in Eq.~\eqref{eq:q_target} uses
$V'(s_{t+1})$ and does not require Q-value estimates for actions predicted by
the current actor. This anchors critic learning to the replay data.
Moreover, $V'(s_{t+1})$ is evaluated directly from the cached state
representation, so each critic update avoids invoking the VLA to predict a new
action chunk at $s_{t+1}$. Therefore, changes to the actor do not require
regenerating target actions, removing an expensive VLA forward pass from each
critic update and substantially improving training efficiency as shown in Fig.~\ref{fig:ablation_curves}.

\subsection{Replay-Anchored One-Step Policy Improvement}
Using the off-policy critic learned above, we perform Q-based
advantage-weighted regression over recorded behavior chunks. Flow-based policies commonly generate an action chunk by integrating a learned vector
field from an initial Gaussian latent over multiple solver
steps~\cite{lipman2023flow}. We retain the original flow policy and simply use a single solver step for both rollout and online policy learning, without introducing a separate one-step actor or an additional distillation stage. During rollout, given state $s_t$, the rollout actor
samples $z_t\sim\mathcal N(0,I)$ once. Gaussian exploration noise $\epsilon_t\sim\mathcal N(0,\Sigma)$ is added to the one-step flow output,
yielding $\mathbf a_t=F_{\bar\theta}(s_t,z_t)+\epsilon_t$. The sampled $z_t$
is stored with the resulting transition.

During learning, $z_t$ is retrieved from replay rather than resampled. The
learner actor uses this stored latent to produce the current noise-free
prediction
$\mathbf a_{t,\theta}^{\mathrm{clean}}=F_\theta(s_t,z_t)$.
Resampling $z_t$ could select a different stochastic branch and misalign the
actor prediction with the recorded behavior chunk. Because $z_t$ is drawn from
a fixed Gaussian base distribution rather than generated by the rollout
policy, it remains a policy-independent conditioning variable as the actor
changes. This allows the same transition to be reused across actor updates
without storing policy-dependent intermediate flow states.
Let $\mathbf e_t=\mathbf a_t^b-\mathbf a_{t,\theta}^{\mathrm{clean}}$ denote
the residual between the behavior chunk and the current actor
prediction. Conditioned on the stored latent $z_t$, the current actor assigns
\begin{equation}
\log p_\theta(\mathbf a_t^b\mid s_t,z_t)
=-\tfrac{1}{2}\mathbf e_t^\top\Sigma^{-1}\mathbf e_t
-\tfrac{1}{2}\log\det(2\pi\Sigma).
\label{eq:conditional_density}
\end{equation}
Here, $\mathbf a_t^b$ is the final recorded behavior chunk: it is the rollout
action for autonomous transitions and the human-corrected action for
intervention transitions. In both cases, Eq.~\eqref{eq:conditional_density}
trains the current actor toward $\mathbf a_t^b$ under the stored latent.

Specifically, we compute the replay advantage and its clipped exponential
weight as
\begin{align}
A_t&=\min_{i\in\{1,2\}}Q_i'(s_t,\mathbf a_t^b)-V(s_t),\\
w_t&=\exp\!\left(\operatorname{clip}(A_t/\beta,-c,c)\right).
\label{eq:awac_weight}
\end{align}
In practice, we set $\beta=0.5$ and $c=3$, where $\beta$ controls the
strength of advantage weighting and $c$ bounds the log weight. We optimize the
actor with
\begin{equation}
\mathcal L_{\mathrm{actor}}=-
\mathbb E_{(s_t,\mathbf a_t^b,z_t)\sim\mathcal D}
\left[w_t\log p_\theta(\mathbf a_t^b\mid s_t,z_t)\right].
\label{eq:actor_loss}
\end{equation}
Higher-advantage chunks therefore receive larger likelihood updates,
as in AWAC~\cite{nair2020awac}. Since the likelihood is evaluated on recorded
behavior chunks, actor optimization remains anchored to replayed actions, with
$w_t$ treated as fixed.

The one-step mapping requires only one action-expert evaluation for both action
generation and likelihood computation. Unlike methods that distill a
multi-step flow policy into a separate one-step actor~\cite{park2025fql}, our
actor learns directly from the aligned replay tuples, requiring neither a
multi-step teacher nor an additional distillation stage. Therefore, it avoids unrolling a multi-step flow solver during replay updates, keeping both rollout
and actor learning computationally stable and efficient.

\section{EXPERIMENTS}
\label{sec:experiments}

\subsection{Experimental Setup}
\subsubsection{Tasks}
\begin{figure}[t]
    \centering
    \includegraphics[width=\columnwidth]{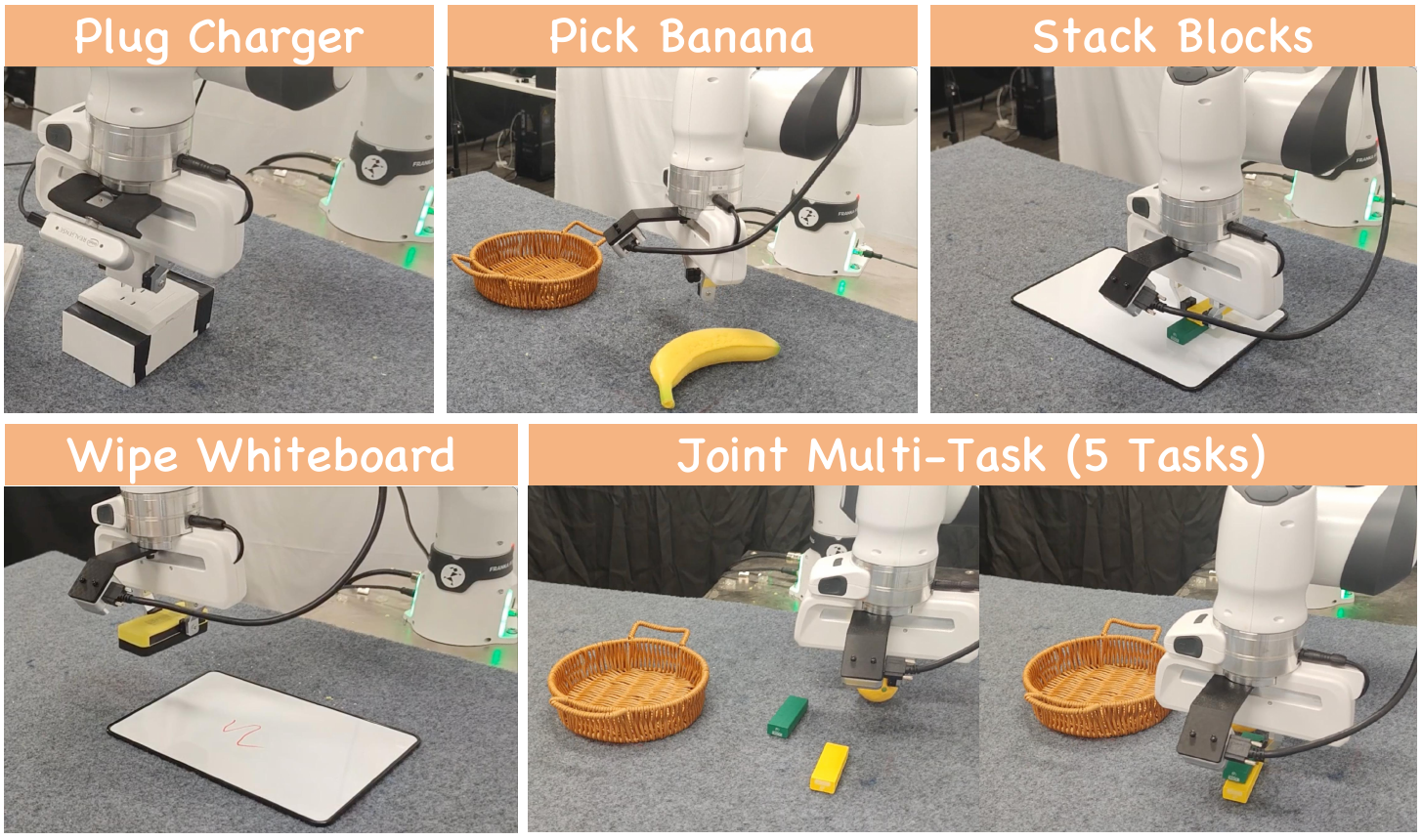}
    \caption{Real-world tasks: four single-task settings covering precision
    insertion, pick-and-place, stacking, and contact-rich wiping, and a joint multi-task
    setting with five language-conditioned pick-and-place and stacking tasks.}
    \label{fig:experiment_tasks}
\end{figure}

We evaluate RAPolicy in five real-world settings: four single-task settings and
one joint multi-task setting. These settings span precision
manipulation, contact-rich interaction, visual understanding, and
language-conditioned multi-task control, as illustrated in
Fig.~\ref{fig:experiment_tasks}.

\paragraph{Plug Charger.}
This precision insertion task uses the language prompt
\emph{``plug the charger into the socket''}. The robot must align the charger
plug with the socket and complete the insertion.

\paragraph{Pick Banana.}
This pick-and-place task uses the language prompt
\emph{``put the banana into the basket''}. The robot must localize and grasp
the banana, transport it without dropping it, and release it inside the basket.

\paragraph{Stack Blocks.}
This precision stacking task uses the language prompt
\emph{``stack the yellow block on top of the green block''}. The robot must
grasp the yellow block, align it above the green block, and release it to form
a stable stack.

\paragraph{Wipe Whiteboard.}
This contact-rich and visually grounded manipulation task uses the language prompt
\emph{``pick up the whiteboard eraser and erase the red writing from the
whiteboard''}. The robot must grasp the eraser, locate the writing, and
maintain contact with the board while wiping.

\begin{figure*}[t]
    \centering
    \includegraphics[width=\textwidth]{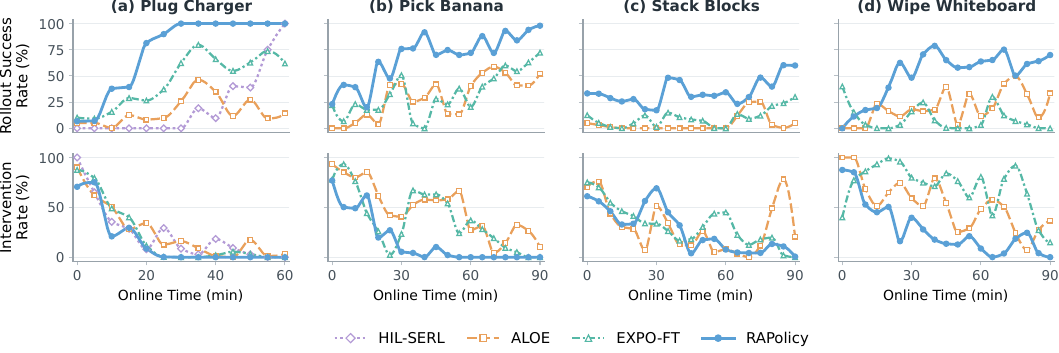}
    \caption{Single-task rollout success rates (top) and intervention rates
    (bottom) versus online training time. RAPolicy improves task success more
    rapidly while requiring fewer human interventions.}
    \label{fig:rollout_success_wallclock}
\end{figure*}

\begin{table*}[t]
    \centering
    \caption{Single-task evaluation success rates (\%) for initial SFT and final
    online policies, evaluated over 20 trials per task.}
    \label{tab:single_task_evaluation}
    \setlength{\tabcolsep}{8pt}
    \renewcommand{\arraystretch}{1.1}
    \begin{tabular}{lccccc}
        \toprule
        Task & SFT ($\pi_{0.5}$) & HIL-SERL & ALOE & EXPO-FT & RAPolicy \\
        \midrule
        Plug Charger & 5\% (1/20) & \textbf{100\% (20/20)} & 30\% (6/20) &
        80\% (16/20) & \textbf{100\% (20/20)} \\
        Pick Banana & 15\% (3/20) & -- & 55\% (11/20) & 80\% (16/20) &
        \textbf{95\% (19/20)} \\
        Stack Blocks & 0\% (0/20) & -- & 5\% (1/20) & 30\% (6/20) &
        \textbf{70\% (14/20)} \\
        Wipe Whiteboard & 10\% (2/20) & -- & 50\% (10/20) & 10\% (2/20) &
        \textbf{80\% (16/20)} \\
        \bottomrule
    \end{tabular}
\end{table*}

\paragraph{Joint Multi-Task.}
This language-conditioned setting trains one shared policy to execute 5
tasks in the same scene. The prompts are \emph{``Put the lemon into the
basket.''}, \emph{``Put the yellow block into the basket.''}, \emph{``Put the
green block into the basket.''}, \emph{``Stack the yellow block on top of the
green block.''}, and \emph{``Stack the green block on top of the yellow
block.''}. The setting combines pick-and-place and precision stacking while
requiring the policy to distinguish the commanded object and goal relation.

\subsubsection{Comparisons}
For single-task adaptation, we compare RAPolicy with ALOE~\cite{yang2026aloe}, EXPO-FT~\cite{dong2026expoft}, and HIL-SERL~\cite{luo2024hilserl}.
ALOE uses action-level off-policy evaluation and
chunk-level TD learning for batch-online VLA post-training.
EXPO-FT extends EXPO to fine-tune a pretrained VLA with
temporally extended actions and human intervention.
HIL-SERL trains a compact policy from scratch with
off-policy RL and human intervention. We found that, except for {\it Plug Charger},
the tasks involve broader visual distributions and longer horizons, making it
difficult for HIL-SERL to learn an effective policy within the limited online
training budget. We therefore report HIL-SERL results only for {\it Plug Charger}.
For the joint multi-task setting, we compare RAPolicy with
HG-DAgger~\cite{ross2011dagger,hoque2022thriftydagger,kelly2019hgdagger}, an interactive imitation-learning method
that updates the policy from human corrections without reward-based policy
optimization. The other RL baselines are not extended to this setting because
their single-task results already show slow adaptation within the available
online training budget, making evaluation in the more demanding joint
multi-task setting impractical under the same budget.
Notably, we implement ALOE asynchronously; consequently, rollout proceeds
continuously throughout online training for all methods, ensuring a fair
comparison in online data collection. For methods that require predicting next
actions, we cache policy-generated actions to accelerate critic
training and refresh the cache after each policy update.

\subsubsection{Implementation Details}
We conduct all experiments on a Franka Research 3. Each task uses two RGB cameras, comprising a wrist-mounted view and a third-person view.
The action-chunk horizon is $H=5$ for {\it Plug Charger} and $H=10$ for all other tasks.
All experiments use a binary sparse reward, with a reward of 10 for successful task
completion and 0 otherwise.

In the single-task setting, we first perform 10-demos SFT and then run online RL. 
During online process, we don't use the offline data for buffer initialization, so this setting primarily evaluates the method's online adaptation efficiency. In the joint multi-task setting, we collect 30
demonstrations per task, for a total of 150, and use all of them to
train a shared SFT policy before online RL. Here, the offline data
are used for buffer initialization, and the setting evaluates the stability, effectiveness, and speed of joint online improvement.

We use a batch size of 112 and run the system in a distributed configuration. A local control computer handles only robot control, while a remote server with 8 NVIDIA RTX 3090 GPUs runs rollout (1 GPU) and training (7 GPUs).

\subsubsection{Evaluation Metrics}
We evaluate online learning using rollout success rate, intervention rate,
and robot-data efficiency, and evaluate each final policy over 20 trials. For time interval $i$, let $N_i$ denote the number
of rollout episodes recorded in that interval, $N_i^{\mathrm{succ}}$ the number
of successful episodes without human intervention, and
$N_i^{\mathrm{int}}$ the number of episodes with at least one intervention. We
compute
\begin{equation}
\mathrm{SR}_i=\frac{N_i^{\mathrm{succ}}}{N_i},\qquad
\mathrm{IR}_i=\frac{N_i^{\mathrm{int}}}{N_i}.
\end{equation}
We report these metrics against elapsed online time and cumulative robot interaction time in minutes. Higher success under comparable time or data budgets indicates greater adaptation efficiency, while retaining acquired performance during continued learning indicates stability.

\begin{figure}[t]
    \centering
    \includegraphics[width=\columnwidth]{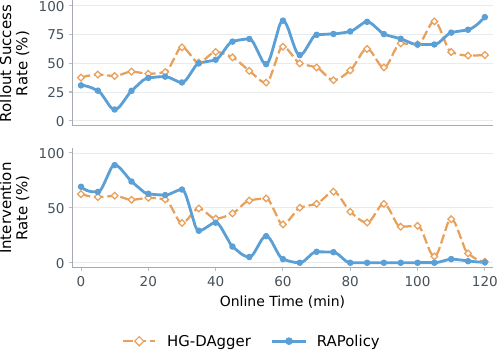}
    \caption{Joint five-task rollout success rates (top) and intervention
    rates (bottom) versus online training time. RAPolicy achieves greater
    improvement in the shared language-conditioned policy while requiring
    fewer human interventions than HG-DAgger, reaching approximately 90\%
    rollout success rate after 120 minutes of training.}
    \label{fig:multitask_training}
\end{figure}

\subsection{Single-Task Online Adaptation}
Fig.~\ref{fig:rollout_success_wallclock} compares the single-task online
learning curves. On the fine-grained {\it Plug Charger} task, RAPolicy converges within 30 minutes and
maintains 100\% rollout success for the remainder of training. HIL-SERL
attains the same rollout success only at the end of its 60-minute run, whereas
EXPO-FT and ALOE remain at 61.9\% and 14.3\%, respectively. On
{\it Pick Banana} task, RAPolicy reaches 91.7\% at 40 minutes and,
despite subsequent fluctuations, finishes at 98\% success without
intervention; EXPO-FT and ALOE finish at 72.2\% and 52.2\%. For the more
challenging tasks, RAPolicy finishes at 60\% success on the fine-grained
\textit{Stack Blocks} task and 70\% on the contact-rich
\textit{Wipe Whiteboard} task, whereas the best baselines reach only 30\% and 33.3\%, respectively.
Overall, RAPolicy achieves substantially faster and stable policy improvement and greater
online training efficiency than the baselines while requiring fewer human
interventions.

Additionally, we evaluate the final policies obtained after online learning
over 20 trials. For each task, ALOE, EXPO-FT, and RAPolicy initialize online learning from the same $\pi_{0.5}$ model fine-tuned on 10 demonstrations. The results are reported in Table~\ref{tab:single_task_evaluation}. The initial SFT policies achieve success rates of only 5\%, 15\%, 0\%, and 10\% on the four tasks, respectively. Under the same online RL time budget, RAPolicy reaches 100\%, 95\%, 70\%, and 80\%, improving over SFT by 95, 80, 70, and 70 percentage points. RAPolicy matches HIL-SERL on Plug Charger and exceeds the strongest evaluated baseline by 15, 40, and 30 points on the remaining three tasks.
Together, the rollout curves and final evaluations show that RAPolicy adapts
efficiently from weak initial policies across precision, long-horizon, and
contact-rich tasks.

\subsection{Joint Multi-Task Online Post-Training}
\label{sec:multitask}

Fig.~\ref{fig:multitask_training} compares the joint multi-task online learning curves. Starting from the same multi-task $\pi_{0.5}$ SFT model fine-tuned on 150 demonstrations (30 per task), RAPolicy reaches approximately 90\% rollout success over 120 minutes of online training. In contrast, HG-DAgger exhibits larger fluctuations and finishes at 57.1\%. In short, RAPolicy achieves greater improvement in the shared policy while requiring fewer human interventions than HG-DAgger.

\begin{table}[t]
    \centering
    \caption{Joint multi-task evaluation success rates (\%) for initial SFT and final online policies, evaluated over 10 trials per task.}
    \label{tab:multitask_evaluation}
    \small
    \setlength{\tabcolsep}{3pt}
    \renewcommand{\arraystretch}{1.1}
    \resizebox{\columnwidth}{!}{%
    \begin{tabular}{lccc}
        \toprule
        Task & SFT ($\pi_{0.5}$) & HG-DAgger & RAPolicy \\
        \midrule
        Lemon (PnP) & 40\% (4/10) & 70\% (7/10) & \textbf{90\% (9/10)} \\
        Yellow Block (PnP) & 90\% (9/10) & 90\% (9/10) & \textbf{100\% (10/10)} \\
        Green Block (PnP) & \textbf{100\% (10/10)} & \textbf{100\% (10/10)} & \textbf{100\% (10/10)} \\
        Y$\rightarrow$ G (Stack) & 20\% (2/10) & 30\% (3/10) & \textbf{90\% (9/10)} \\
        G$\rightarrow$ Y (Stack) & 10\% (1/10) & \textbf{60\% (6/10)} & \textbf{60\% (6/10)} \\
        \midrule
        Overall & 52\% (26/50) & 70\% (35/50) & \textbf{88\% (44/50)} \\
        \bottomrule
    \end{tabular}
    }
\end{table}

\begin{figure}[t]
  \centering
  \includegraphics[width=\columnwidth]{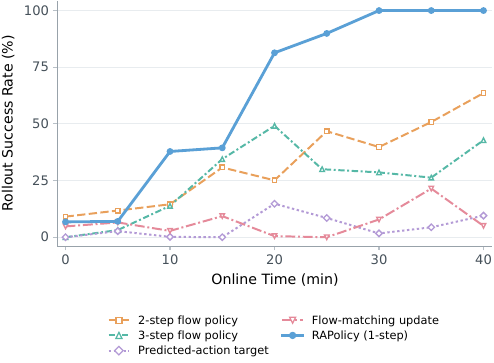}
  \caption{Online ablations on \textit{Plug Charger} task. RAPolicy is compared with variants using two or
  three flow steps, predicted-action critic targets, or flow-matching actor
  updates, with similar numbers of human interventions.}
  \label{fig:ablation_curves}
\end{figure}

\begin{figure*}[t]
    \centering
    \includegraphics[width=\textwidth]{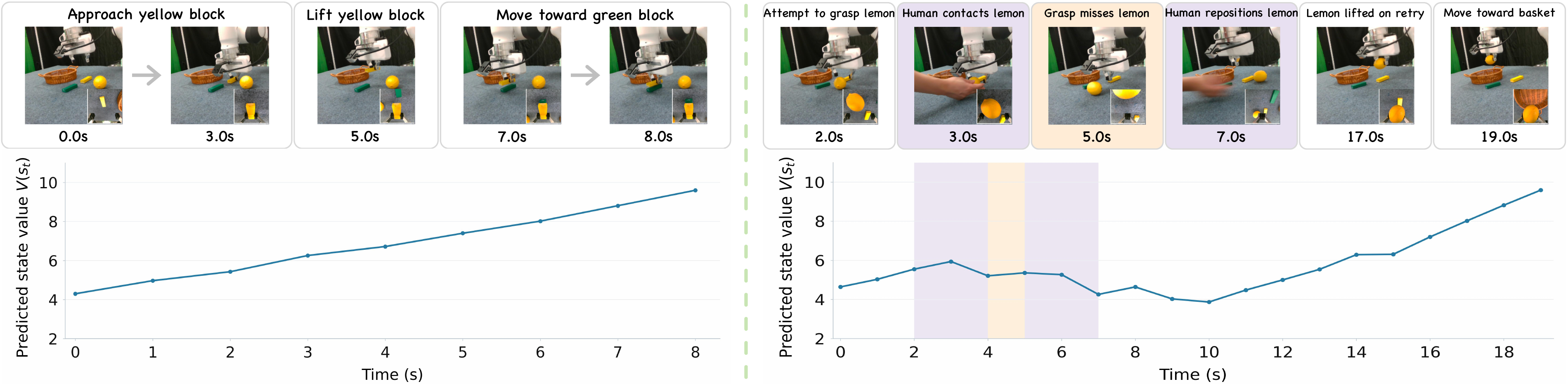}
    \caption{Learned state values $V(s_t)$ along two rollout trajectories.
    Left: values rise as the robot lifts the yellow block and moves it toward
    the green block. Right: values decrease after a human displaces the lemon
    before grasping, then recover as the robot grasps it and moves toward the
    basket.}
    \label{fig:value_visualization}
  \end{figure*}

Additionally, we evaluate the final policies obtained after online learning
over 10 trials per task. The results are reported in Table~\ref{tab:multitask_evaluation}. The initial SFT policy achieves an overall success rate of 52\%. Under the same online RL time budget, RAPolicy
reaches 88\%, improving over SFT by 36 percentage points and outperforming
HG-DAgger at 70\% by 18 percentage points. When aggregating the two stacking tasks, RAPolicy achieves the largest gain, improving the combined success rate from 15\% (3/20) to 75\% (15/20). It also improves {\it Pick Lemon} task from 40\% to 90\% and reaches 100\% on both block PnP tasks. Together, the rollout curves and final evaluations show that RAPolicy enables efficient and stable improvement of a shared language-conditioned policy, with the largest gains on capabilities that remain weak after SFT while preserving those that are already reliable.

\subsection{Ablation Study}
\label{sec:ablation}
In this section, we ablate three key design choices of RAPolicy on \textit{Plug
Charger} task, with all comparisons using similar numbers of human interventions.
We then visualize the learned state value $V(s_t)$ during inference.

\paragraph{One-Step vs. Multi-Step Flow Policies}
Fig.~\ref{fig:ablation_curves} compares actors using 1, 2, or 3 flow
steps. RAPolicy directly optimizes the mapping from
the stored rollout latent $z_t$ to the replayed action $\mathbf a_t^b$. With the
one-step policy, the advantage-weighted objective directly supervises this
mapping. Multi-step policies instead obtain the final action by unrolling the
flow solver and backpropagating the endpoint loss through policy-dependent
intermediate action states. Because these intermediate states are regenerated
by the evolving actor and are not directly constrained by replay, the resulting
optimization is less direct and can be more sensitive to errors across the
unrolled steps. Empirically, the one-step policy converges within 30 minutes and
maintains 100\% rollout success rate, whereas the two- and three-step variants reach
63.6\% and 42.8\% at 40 minutes, respectively. These results suggest that the
direct one-step mapping provides more stable and effective replay-based actor supervision
in our setting.

\paragraph{Replay-Anchored Critic Learning vs. Predicted-Action Bootstrapping}
RAPolicy bootstraps with $V'(s_{t+1})$, which is fitted only to Q-values of replayed
action chunks. We replace this target with
$\min_i Q_i'(s_{t+1},\hat{\mathbf a}_{t+1})$, where the current
actor predicts $\hat{\mathbf a}_{t+1}$. In asynchronous online learning, this
predicted action changes whenever the actor is updated and may be weakly
supported by replay. Its Q-value can therefore introduce policy-dependent
estimation error into the TD target. Replay anchoring removes this dependence
and keeps value learning grounded in recorded behavior. Although we cache
$\hat{\mathbf a}_{t+1}$ to substantially accelerate critic training, the cache
must be refreshed after each policy update. Consequently, each update group requires more than $3\times$ time required by RAPolicy. As shown in
Fig.~\ref{fig:ablation_curves}, the replay-anchored variant converges within 30
minutes, whereas the predicted-action variant remains at 10\% at 40 minutes. Moreover,
even after 120 minutes of continuous online training, the predicted-action
variant reaches only approximately 40\% rollout success rate. These results
demonstrate that replay anchoring both reduces critic-update cost and provides
more reliable value targets for stable online policy improvement.

\paragraph{Replay-Anchored Actor Updates vs. Flow-Matching Updates}
We replace the stored-latent conditional-likelihood objective in
Eq.~\eqref{eq:actor_loss} with the advantage-weighted flow-matching objective
used by ALOE~\cite{yang2026aloe}.
For each replayed action, flow matching samples a new noise vector and
supervises a path connecting it to the action, without preserving the
noise--action pairing from rollout. RAPolicy instead retains this pairing: the
stored $z_t$ identifies the input that produced the recorded behavior, and the
advantage-weighted residual $\mathbf a_t^b-F_\theta(s_t,z_t)$ directly adjusts
the corresponding one-step output. This directly applies advantage-weighted
supervision to the one-step action mapping used during rollout. As shown in
Fig.~\ref{fig:ablation_curves}, RAPolicy converges within 30 minutes, whereas the flow-matching variant's rollout success
rate still fluctuates around 10\% after 40 minutes of training. These results
support the effectiveness of stored-latent endpoint
supervision for online improvement of our one-step policy.

\paragraph{Learned Value Visualization}
In Fig.~\ref{fig:value_visualization}, we select two representative rollout
trajectories to visualize the state values $V(s_t)$ learned by the final policy.
In Fig.~\ref{fig:value_visualization} (left), the value rises steadily from $4.3$ to
$9.6$ as the robot approaches, lifts, and moves the yellow block toward the
green block. In Fig.~\ref{fig:value_visualization} (right), the value decreases after
the lemon is deliberately moved away by a human just before the robot grasps
it, introducing an external disturbance to test policy robustness. It then
recovers as the robot reapproaches and successfully lifts the displaced lemon
before moving it toward the basket. Despite being trained with only binary task rewards, the replay-anchored critic assigns higher values to task progress and lower values to temporary setbacks, while capturing the subsequent recovery from the human-induced disturbance. The learned $V(s_t)$ therefore provides a progress-sensitive baseline for the
advantage weights used to improve the actor from replayed action chunks.

\section{CONCLUSION}
\label{sec:conclusion}

We present RAPolicy, a framework for real-world online VLA post-training that
combines asynchronous execution with replay-anchored critic and actor learning.
Its critic constructs chunk-level value targets without predicting next
actions, while its one-step actor improves recorded behavior through
conditional likelihood using stored rollout latents. Experiments demonstrate
effective adaptation and improvement of a shared
policy across five language-conditioned tasks, with fewer human interventions
during online learning. Ablations further support the one-step policy design
and the effectiveness of anchoring both value targets and actor supervision to
replay. Together, these findings support replay-anchored learning as a practical
approach to efficient online improvement of VLA policies in the real world.

\balance
\bibliographystyle{IEEEtranBST/IEEEtran}
\bibliography{ieeeconf/references}

\end{document}